\documentclass{article}

\usepackage[preprint]{neurips_2025}

\usepackage[utf8]{inputenc} % allow utf-8 input
\usepackage[T1]{fontenc}    % use 8-bit T1 fonts
\usepackage{hyperref}       % hyperlinks
\usepackage{url}            % simple URL typesetting
\usepackage{booktabs}       % professional-quality tables
\usepackage{amsfonts}       % blackboard math symbols
\usepackage{nicefrac}       % compact symbols for 1/2, etc.
\usepackage{microtype}      % microtypography
\usepackage{xcolor}         % colors
\usepackage{graphicx}
\usepackage{amsmath}

\title{MathBode: Measuring the Stability of LLM Reasoning using Frequency Response}
\author{
  Charles L. Wang \\
  Department of Computer Science \\
  Columbia University \\
  \texttt{charles.w@columbia.edu}
}

\begin{document}

\maketitle

\begin{abstract}
We present \textbf{MathBode}, a \emph{dynamic diagnostic} for mathematical reasoning in large language models (LLMs). Instead of one-shot accuracy, MathBode treats each parametric problem as a system: we drive a single parameter sinusoidally and fit first-harmonic responses of model outputs and exact solutions. This yields interpretable, frequency-resolved metrics---\emph{gain} (amplitude tracking) and \emph{phase} (lag)---that form Bode-style fingerprints. Across five closed-form families (linear solve, ratio/saturation, compound interest, $2{\times}2$ linear systems, similar triangles), the diagnostic surfaces systematic \emph{low-pass} behavior and growing phase lag that accuracy alone obscures. We compare several models against a symbolic baseline that calibrates the instrument ($G \!\approx\! 1$, $\phi \!\approx\! 0$). Results separate frontier from mid-tier models on dynamics, providing a compact, reproducible protocol that complements standard benchmarks with actionable measurements of reasoning fidelity and consistency. We open-source the dataset and code to enable further research and adoption. \hspace{0.1in}
{\small
\noindent
\href{https://github.com/charleslwang/MathBode-Eval}{\raisebox{-0.2ex}{\includegraphics[height=1.05em]{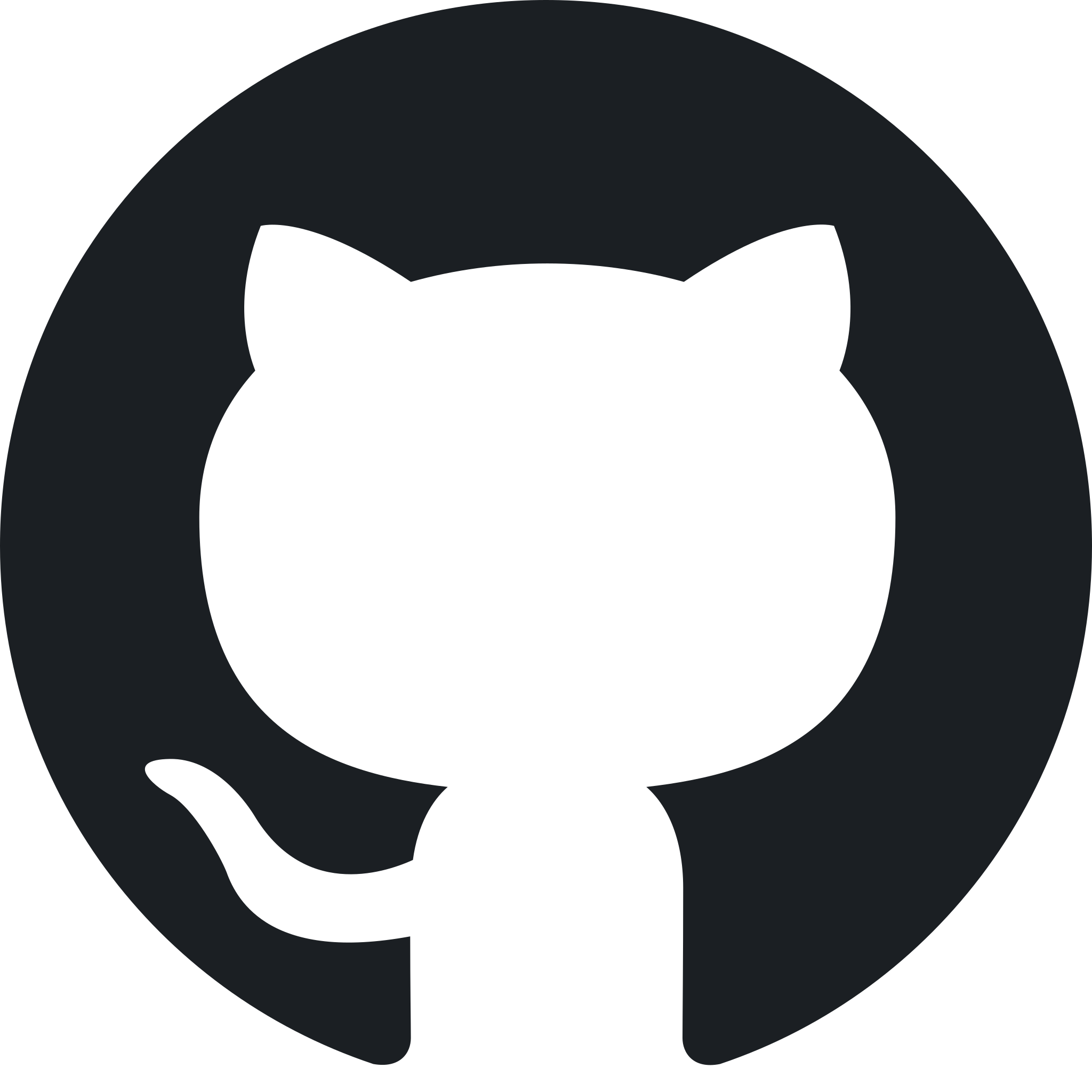}}\; Code}
\quad|\quad
\href{https://huggingface.co/datasets/cognitive-metrology-lab/MathBode}{\raisebox{-0.2ex}{\includegraphics[height=1.05em]{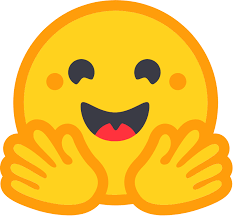}}\; Dataset}
}
\end{abstract}

% =========================
% content.tex  
% =========================

\section{Introduction}

Large language models (LLMs) now score highly on math benchmarks, but final–answer accuracy obscures \emph{how} they reason and whether behavior is stable under controlled changes. We propose a \emph{dynamic} evaluation: treat each parametric problem as a system, drive one parameter sinusoidally, and summarize the model’s response by \emph{gain} (amplitude tracking) and \emph{phase} (lag) over frequency. \textbf{MathBode} implements this across five closed-form families, fitting first-harmonic responses to produce Bode-style fingerprints that reveal low-pass behavior and growing phase lag even when static accuracy ties. The protocol is simple (short prompts, deterministic decoding) and includes a symbolic baseline to calibrate the instrument (ideal $G\!\approx\!1$, $\phi\!\approx\!0$). We report $G(\omega)$, $|\phi(\omega)|$, mid-band aggregates, residual autocorrelation, and first-harmonic fit quality ($R^2$), providing a complementary lens on reasoning fidelity, consistency, and prompt sensitivity that accuracy alone cannot capture.

\paragraph{Context.}
Progress in mathematical reasoning is typically reported on static, final-answer datasets such as GSM8K and MATH, with domain-tuned systems (e.g., Minerva) pushing scores higher \citep{cobbe2021gsm8k,hendrycks2021math,lewkowycz2022minerva}. Newer suites emphasize expert difficulty and recency—OlympiadBench, Omni-MATH, FrontierMath—yet still follow the one-input/one-answer paradigm \citep{he2024olympiadbench,gao2024omnimath,glazer2024frontiermath}. A parallel thread probes robustness: small semantic edits can flip answers (SVAMP; MATH-Perturb), while sampling strategies like self-consistency improve end accuracy without \emph{measuring} stability \citep{patel2021svamp,huang2025mathperturb,wang2022selfconsistency}. Meta-reasoning probes and repeated-trial consistency likewise show models can be correct once yet unreliable across paraphrases or restarts \citep{zeng2023mrgsm8k}. Together, these observations motivate metrics that capture reliability and invariance, not just correctness.

\paragraph{Why a frequency/phase view?}
Interpretability results suggest a principled bridge to the frequency domain: transformers trained on arithmetic learn sinusoidal/rotational internal codes; modular addition emerges via Fourier-like features and rotations; recent work describes clock-like number embeddings and trigonometric operations \citep{nanda2023grokking,kantamneni2025trigAddition,li2024modadd}. If numeric reasoning is expressed in amplitude and phase, then frequency-response style probing is natural rather than metaphorical. 

\paragraph{What MathBode measures.}
For each family, we generate a parameter trajectory $p_t=p_0+\epsilon\sin(\omega t)$, decode a single numeric line with temperature~0, and fit $\{1,\sin(\omega t),\cos(\omega t)\}$ to both ground truth and model outputs. From the fitted coefficients we recover amplitude and phase and compute
$G(\omega)=\mathrm{amp}(\hat y)/\mathrm{amp}(y^\ast)$ and
$\phi(\omega)=\mathrm{wrap}\bigl(\phi(\hat y)-\phi(y^\ast)\bigr)$.
We sweep $\omega\in\{1,2,4,8,16\}$ (64 steps), optionally vary start phase to assess phase stability, and include a symbolic baseline that realizes the ideal response. The resulting frequency-resolved curves and aggregates expose amplitude fidelity, timing lag, and prompt-surface sensitivity—even when static accuracy saturates or training-data familiarity blurs the line between recall and robust computation. Deterministic decoding and strict numeric parsing ensure we compare numeric sequences, not templates. A generic pattern or echo policy would typically yield incorrect amplitude/timing (non-unity $G$, shifted $\phi$) and elevated residual autocorrelation, even if surface formatting looked consistent.

\section{Benchmark}
\label{sec:method-benchmark}

\textbf{Instrument.}
We probe \emph{dynamic} mathematical reasoning by driving one problem parameter with a sinusoid and fitting first-harmonic responses of model outputs against exact solutions. For a sweep of length $T$ and angular frequency $\omega$ we instantiate prompts with
\[
p_t = p_0 + \epsilon \sin(\omega t + \phi_0), \quad t=1,\ldots,T,
\]
decode deterministically (temperature $0$) to a single numeric line (\texttt{FINAL: <number>}), and parse the model series $\hat y_t$ alongside the exact series $y_t^\ast$. Each series is regressed onto $\{\sin(\omega t),\cos(\omega t),1\}$; from the fitted coefficients $(a,b,c)$ we recover amplitude and phase
\[
\mathrm{amp}(y)=\sqrt{a^2+b^2}, \qquad \phi(y)=\mathrm{atan2}(b,a).
\]
We then report
\[
G(\omega)=\frac{\mathrm{amp}(\hat y)}{\mathrm{amp}(y^\ast)}, \qquad
\phi(\omega)=\mathrm{wrap}_{(-\pi,\pi]}\!\bigl(\phi(\hat y)-\phi(y^\ast)\bigr),
\]
along with first-harmonic $R^2$ (fit quality), residual RMS (normalized), residual ACF(1), and a nonlinearity proxy $H_2/H_1$ from a joint fit at $\omega$ and $2\omega$. A symbolic solver baseline runs through the identical pipeline, providing the ideal reference ($G\!\approx\!1$, $\phi\!\approx\!0$).

Although gain and phase originate in linear systems, we do \emph{not} assume linear time–invariant behavior. The sinusoid is used purely as a controlled probe: we project both exact and model series onto the first harmonic to summarize amplitude fidelity (gain) and timing (phase), while residual diagnostics and $H_2/H_1$ explicitly capture departures from a single-tone (e.g., nonlinearity and memory). Mechanistic findings of sinusoidal/rotational number codes \citep{nanda2023grokking,kantamneni2025trigAddition,li2024modadd} motivate this descriptive frequency lens rather than a modeling assumption.

\textbf{Families.}
We evaluate five closed-form families with fixed domains and three question variants each: \emph{Linear Solve} ($a{=}p$: solve $x$ in $ax{+}b{=}c$), \emph{Ratio Saturation} ($p/(p{+}k)$), \emph{Exponential Interest} ($A(1{+}p)^t$), \emph{Linear System} (solve $x$ in a $2{\times}2$ system with $a{=}p$), and \emph{Similar Triangles} (scaling $s'=s\,p$). Families expose $(p_\text{range}, p_0, \epsilon)$ via code, and inputs are clipped in-range.

\textbf{Frequency grid and phases.}
We choose $T{=}64$ and sweep $\Omega=\{1,2,4,8,16\}$ cycles per 64 steps. To assess phase robustness we use start phases $\{0^\circ,120^\circ,240^\circ\}$. Defaults set $\epsilon$ to roughly $10\%$ of the family’s half-range.

All experiments use temperature $0$ (deterministic decoding) and strict numeric parsing with compliance filtering.
At $\omega{=}16$ (16 cycles over $T{=}64$), the drive approaches the Nyquist limit; small dips in $R^2$ or phase swings can include aliasing artefacts, so we emphasize the mid-band $\{4,8\}$ region for ranking.

\textbf{Why this design?}
Gain and phase isolate amplitude tracking and lag—two core behaviors that final-answer accuracy obscures—while $R^2$ and residual diagnostics validate the first-harmonic approximation and expose structure left unexplained by it. The frequency grid (with tri-phase repeats) yields stability bands rather than single-shot outcomes, and the symbolic baseline calibrates the measurement end-to-end. The result is an inexpensive, reproducible instrument that complements static accuracy with a frequency-domain lens on reasoning fidelity and consistency.

\section{Dataset Details}

% Cardinality by family
\paragraph{Cardinality.}
\textsc{MathBode} contains \textbf{9{,}408 rows per family} and \textbf{47{,}040 rows total} across five families.

\begin{table}[h]
  \centering
  \small
  \caption{\textbf{Dataset rows by family.}}
  \label{tab:cardinality-by-family}
  \begin{tabular}{l r}
    \toprule
    \textbf{Family} & \textbf{Rows} \\
    \midrule
    Exponential Interest & 9{,}408 \\
    Linear Solve         & 9{,}408 \\
    Linear System        & 9{,}408 \\
    Ratio Saturation     & 9{,}408 \\
    Similar Triangles    & 9{,}408 \\
    \midrule
    \textbf{Total}       & \textbf{47{,}040} \\
    \bottomrule
  \end{tabular}
\end{table}

\renewcommand{\arraystretch}{1.15}
\begin{table}[h]
  \centering
  \small
  \begin{tabular}{p{0.22\linewidth} p{0.14\linewidth} p{0.54\linewidth}}
    \toprule
    \textbf{Attribute} & \textbf{Type} & \textbf{Description} \\
    \midrule
    \texttt{family} & string &
      One of \{\textit{exponential\_interest}, \textit{linear\_solve}, \textit{linear\_system}, \textit{ratio\_saturation}, \textit{similar\_triangles}\}. \\
    \texttt{question\_id} & int &
      Variant index within a family. \\
    \texttt{signal\_type} & string &
      Drive label: \{\textit{sinusoid}, \textit{chirp}, \textit{step}\}. \\
    \texttt{amplitude\_scale} & float &
      Relative amplitude (e.g., 0.5, 1.0, 2.5). \\
    \texttt{frequency\_cycles} & float &
      Frequency label (cycles per 64 steps). \\
    \texttt{phase\_deg} & float &
      Start phase (degrees). \\
    \texttt{time\_step} & int &
      Index within the rendered sequence. \\
    \texttt{p\_value} & float &
      Concrete parameter value used to render the prompt. \\
    \texttt{prompt} & string &
      Fully-rendered natural-language question for the instance. \\
    \texttt{ground\_truth} & float &
      Exact numerical answer. \\
    \bottomrule
  \end{tabular}
  \vspace{-0.5em}
\end{table}

\section{Evaluation}
\label{sec:evaluation}

\paragraph{Scores.}
For each family and frequency we compute $G(\omega)=\mathrm{amp}(\hat y)/\mathrm{amp}(y^\ast)$ and $\phi(\omega)=\mathrm{wrap}\!\bigl(\phi(\hat y)-\phi(y^\ast)\bigr)$ from the first-harmonic fit. \textbf{MB-Core} aggregates mid-band $\{4,8\}$ deviations via a normalized combination of $|G{-}1|$ and $|\phi|$ across families. \textbf{MB-Plus} applies multiplicative down-weights derived from first-harmonic $R^2$, residual RMS/ACF(1), and $H_2/H_1$, penalizing responses that are poorly explained or exhibit nonlinear distortion. (Implementation details and ranges are in code; the same normalization is used for all models.)

\paragraph{Why these views?}
Final-answer accuracy hides \emph{how} a model tracks controlled variation. We therefore summarize each family’s response along four complementary axes: \textbf{(i) gain} (amplitude tracking), \textbf{(ii) phase error} (timing/lag), \textbf{(iii) residual autocorrelation} ACF(1) (leftover temporal structure not captured by the first harmonic), and \textbf{(iv) first-harmonic fit quality} $R^2$. Together these expose low-pass behavior, timing slippage, and prompt-surface sensitivity even when accuracy ties. Additional diagnostics (H2/H1 nonlinearity, compliance, phase-stability across start phases) appear in the appendix.

\begin{figure}[h]
  \centering
  \includegraphics[width=\linewidth]{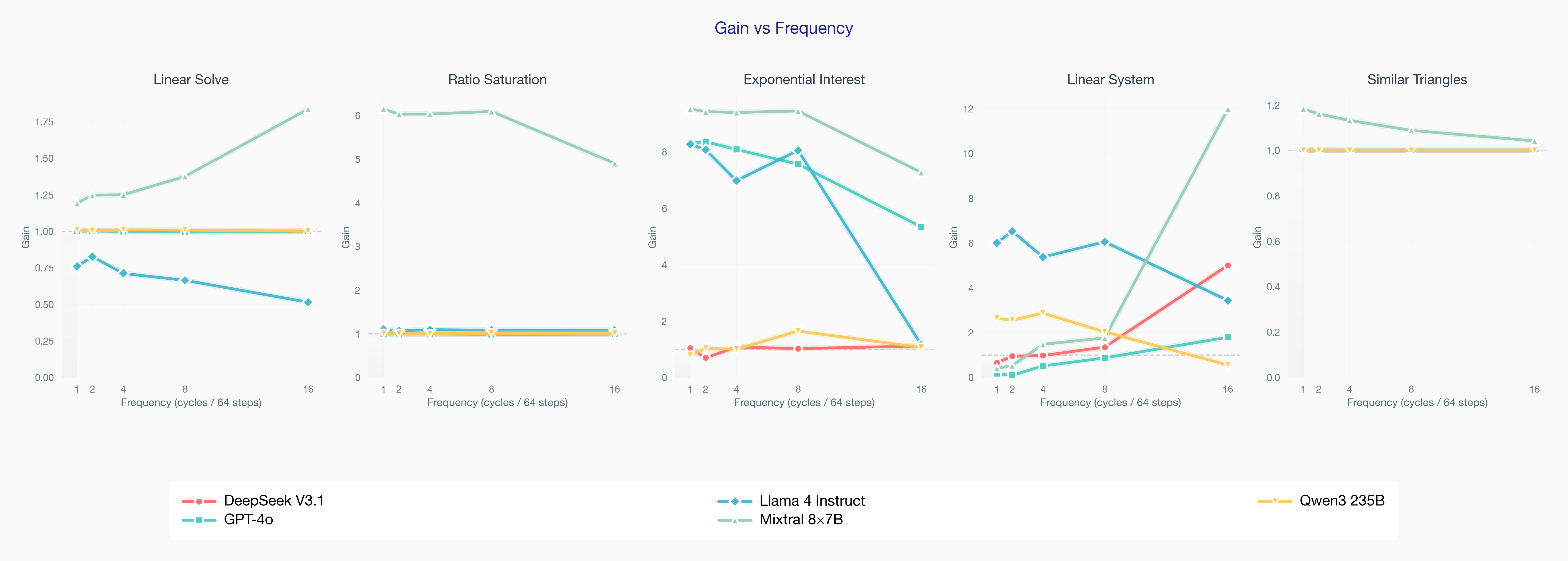}
  \caption{\textbf{Gain vs.\ frequency.} Panels are families; curves overlay models (unity $G{=}1$ dashed). Mid-band (\{4,8\}) deviations indicate under/over-reaction despite identical ground truth.}
  \label{fig:gain}
\end{figure}

\paragraph{Takeaway (Gain).}
Most models are \emph{low-pass}: gain declines with frequency in \emph{Linear Solve} and \emph{Exponential Interest}; \emph{Similar Triangles} stays near $G{\approx}1$ (instrument check). \emph{Linear System} amplifies between-model differences.

\begin{figure}[h]
  \centering
  \includegraphics[width=\linewidth]{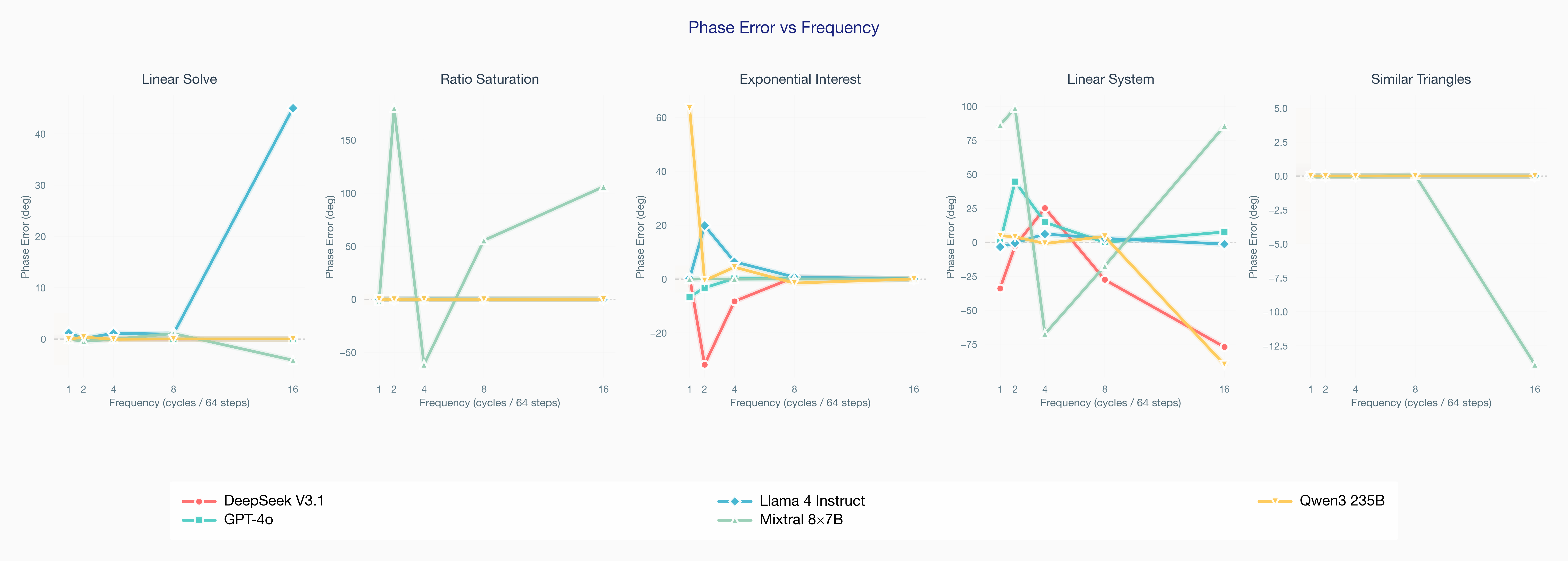}
  \caption{\textbf{Phase error vs.\ frequency.} Signed model–truth phase (rad), wrapped to $(-\pi,\pi]$; $0^\circ$ implies perfect timing.}
  \label{fig:phase}
\end{figure}

\paragraph{Takeaway (Phase).}
Phase lag typically grows with frequency (delayed tracking). Closed-form proportional families (e.g., \emph{Similar Triangles}) remain near $0^\circ$; \emph{Linear System} shows the largest swings (coupling sensitivity).

\begin{figure}[h]
  \centering
  \includegraphics[width=\linewidth]{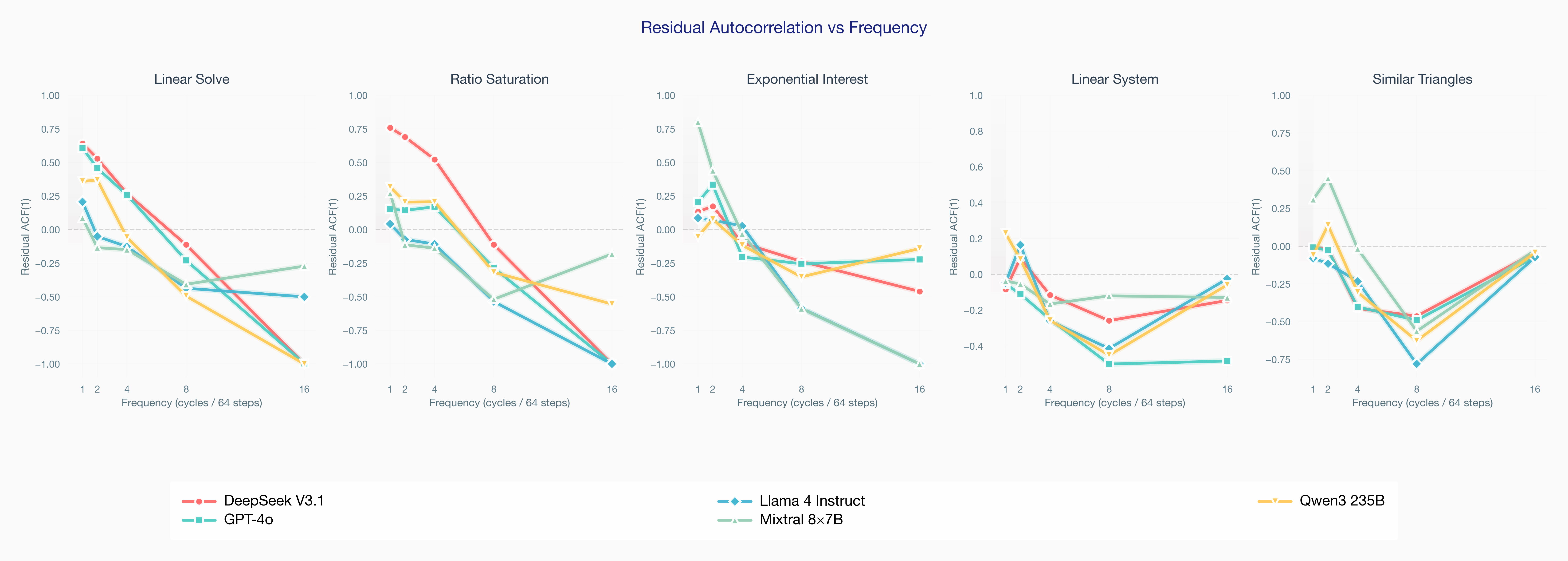}
  \caption{\textbf{Residual ACF(1) vs.\ frequency.} Near-zero ACF(1) means little temporal structure remains after the harmonic fit; negative values align with alternating over/undershoots at higher frequencies.}
  \label{fig:acf}
\end{figure}

\paragraph{Takeaway (Residuals).}
Residual ACF(1) trends toward $0$ or negative with frequency, indicating the first harmonic explains most structure and that remaining errors alternate rather than drift. Residual RMS and H2/H1 curves are provided in the appendix.

\begin{figure}[h]
  \centering
  \includegraphics[width=\linewidth]{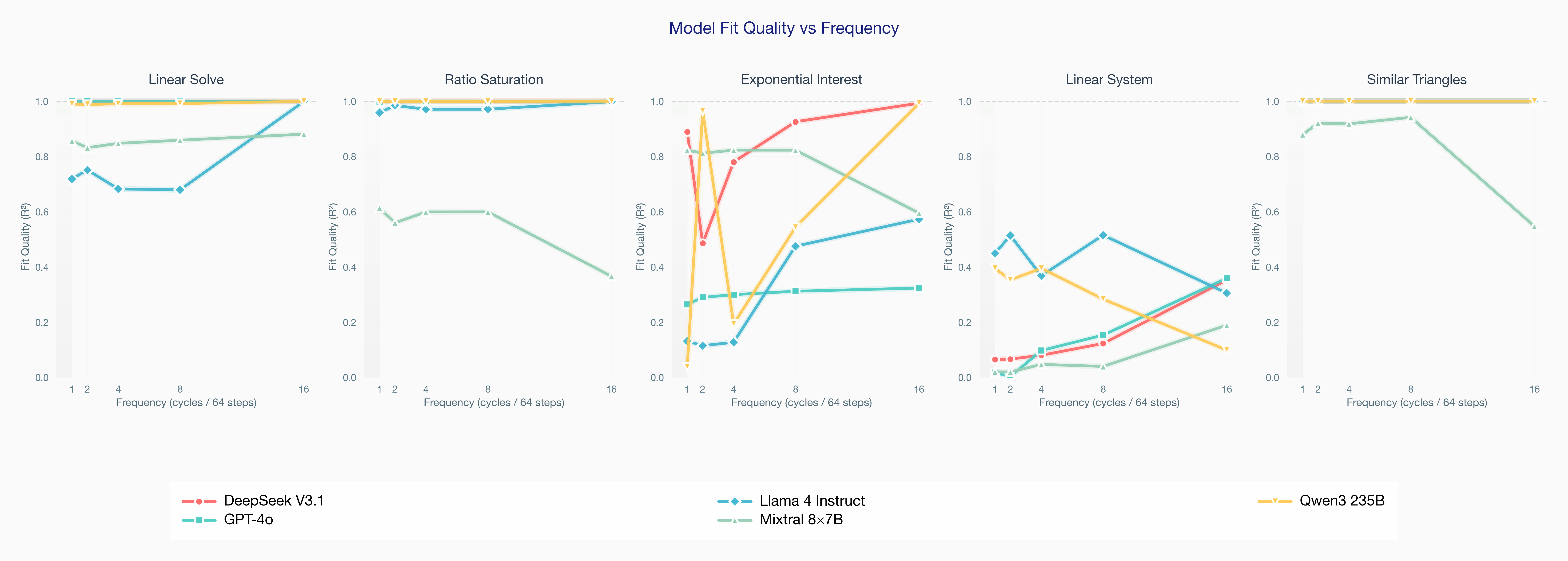}
  \caption{\textbf{First-harmonic fit quality ($R^2$) vs.\ frequency.} High $R^2$ validates a single-sinusoid description; dips signal nonlinear distortion or prompt-surface effects.}
  \label{fig:r2}
\end{figure}

\paragraph{Takeaway ($R^2$).}
$R^2$ is near $1$ for \emph{Similar Triangles} and in the mid-band elsewhere; drops in \emph{Exponential Interest} and \emph{Linear System} co-locate with the largest gain/phase deviations, pointing to emergent nonlinearities rather than random noise. 

\begin{table}[h]
  \centering
  \small
  \setlength{\tabcolsep}{5pt}
  \caption{Overall MathBode scores. MB-Core aggregates mid-band gain/phase deviations; MB-Plus additionally downweights responses with poor fit quality ($R^2$), high residual structure (RMS/ACF), or nonlinearity ($H_2/H_1$). DeepSeek V3.1 leads overall on both MB-Core and MB-Plus.}
  \label{tab:overall-mathbode}
  \begin{tabular}{lcc}
    \toprule
    \textbf{Model} & \textbf{MB-Core} & \textbf{MB-Plus} \\
    \midrule
    \textbf{DeepSeek V3.1}      & \textbf{0.834} & \textbf{0.656} \\
    Qwen3 235B Instruct          & 0.782 & 0.576 \\
    GPT-4o                       & 0.778 & 0.566 \\
    Llama 4 Instruct             & 0.644 & 0.433 \\
    Mixtral 8×7B                 & 0.360 & 0.281 \\
    \bottomrule
  \end{tabular}
\end{table}

\begin{table}[h]
  \centering
  \scriptsize
  \setlength{\tabcolsep}{5pt}
  \caption{Per-family MB-Core (mean mid-band performance).}
  \label{tab:per-family-core}
  \begin{tabular}{lccccc}
    \toprule
    \textbf{Model} & \textbf{Exponential Interest} & \textbf{Linear Solve} & \textbf{Linear System} & \textbf{Ratio Saturation} & \textbf{Similar Triangles} \\
    \midrule
    \textbf{DeepSeek V3.1}       & \textbf{0.848} & \textbf{0.995} & 0.331 & \textbf{0.997} & \textbf{1.000} \\
    GPT-4o                        & 0.497 & 0.993 & 0.418 & 0.980 & \textbf{1.000} \\
    Llama 4 Instruct              & 0.461 & 0.489 & 0.450 & 0.821 & \textbf{1.000} \\
    Mixtral 8×7B                  & 0.500 & 0.494 & 0.029 & 0.000 & 0.779 \\
    Qwen3 235B Instruct           & 0.467 & 0.982 & \textbf{0.471} & 0.990 & \textbf{1.000} \\
    \bottomrule
  \end{tabular}
\end{table}

\section{Conclusion.}
MathBode reframes mathematical evaluation as a dynamic, frequency–domain probe, yielding interpretable gain/phase curves rather than only final answers, moving evaluations towards more reliable mathematical reasoning. Across five closed-form families, models consistently exhibit low-pass behavior and growing phase lag, while the symbolic baseline and our MB-Core/MB-Plus scores summarize these dynamics in a comparable and robust way. The results indicate that strong static accuracy can mask systematic amplitude and timing errors that degrade stability and consistency of reasoning. Practically, the frequency fingerprints provide a compact diagnostic for model selection and ablation studies, complementing standard benchmarks with measurements that are reproducible and easy to interpret. We release the dataset and reference code to support transparent replication and extension. Our use of a sinusoidal drive is an analytical probe rather than an LTI assumption; MB-Core captures mid-band amplitude/timing fidelity, while MB-Plus incorporates explicit penalties for unexplained structure and nonlinearity.
Limitations include the small number of families and single-tone drives; future work will expand the task set, add richer inputs (chirps, steps), and link frequency fingerprints to internal mechanisms (e.g., attention dynamics, layer-wise delays).

\newpage

% ---- References ----
\bibliographystyle{plainnat}
\bibliography{refs}

%%%%%%%%%%%%%%%%%%%%%%%%%%%%%%%%%%%%%%%%%%%%%%%%%%%%%%%%%%%%

\appendix
\section{Appendix}

% =========================
% Presets
% =========================
\section{Presets}
\label{app:presets}

\begin{table}[h]
  \centering
  \small
  \caption{\textbf{Inference presets.} Tri-phase indicates whether phases \{0,120,240\} are used.}
  \label{tab:presets}
  \begin{tabular}{l l l l c}
    \toprule
    \textbf{Preset} & \textbf{Frequencies} & \textbf{Phases} & \textbf{Tri-phase coverage} & \textbf{$K$ (base keys/family)} \\
    \midrule
    \textbf{SMOKE}     & \{4, 8\}                 & \{0\}         & none                         & 2 \\
    \textbf{MVP}       & \{4, 8, 16\}             & \{0\}         & none                         & 2 \\
    \textbf{MVP\_PLUS} & \{1, 2, 4, 8, 16\}       & \{0\}*        & only for \{4, 8\}            & 2 \\
    \textbf{FULL}      & \{1, 2, 4, 8, 16\}       & \{0, 120, 240\} & all frequencies            & 2 \\
    \bottomrule
  \end{tabular}

\paragraph{Note*} In \textbf{MVP\_PLUS}, phases \{0,120,240\} are applied only at mid-band frequencies \{4, 8\}; other frequencies use phase \{0\}.
  \vspace{0.25em}
  \raggedright\footnotesize
\end{table}

% =========================
% Parsing
% =========================
\section{Answer Format \& Strict Parsing}
\label{app:format}
Models output \verb|[answer_start] X.YYYYYY [answer_end]| where the payload is a fixed-precision decimal with exactly six places.

\textbf{Parsing.} From the raw response we (i) find the \emph{last complete} \verb|[answer_start] ... [answer_end]| pair, (ii) scan inside for decimal literals (ASCII digits only; no scientific notation, separators, or units), (iii) take the \emph{last} literal found, and (iv) \emph{truncate} to exactly six decimals (pad with zeros if fewer; cut off if more). Non-finite values (NaN/Inf) or missing tags are non-compliant.

\textbf{Compliance.} Rows that pass this pipeline count as compliant; only compliant rows are used for harmonic fitting and residual diagnostics. Non-compliant rows still contribute to compliance statistics.

% =========================
% Figures & Table
% =========================
\section{Figures \& Tables}
\label{app:figs}
% --- Mid-band |G-1| table ---
\begin{table}[h]
  \centering
  \small
  \caption{\textbf{A.1 Mean \(|G{-}1|\) at mid-frequencies (4 \& 8 cycles).} 
  \emph{Lower is better. EI and LS dominate amplitude error; DeepSeek is best on EI gain, while Mixtral is worst on RS.}}
  \label{tab:midband-gain}
  \begin{tabular}{lrrrrr}
    \toprule
    & \textbf{DeepSeek V3.1} & \textbf{GPT-4o} & \textbf{Llama 4 Instruct} & \textbf{Mixtral 8$\times$7B} & \textbf{Qwen3 235B} \\
    \midrule
    Exponential Interest & 0.051 & 6.819 & 6.512 & 8.418 & 0.323 \\
    Linear Solve         & 0.002 & 0.003 & 0.312 & 0.313 & 0.009 \\
    Linear System        & 0.188 & 0.308 & 4.714 & 0.622 & 1.453 \\
    Ratio Saturation     & 0.002 & 0.010 & 0.087 & 5.059 & 0.005 \\
    Similar Triangles    & 0.000 & 0.000 & 0.000 & 0.110 & 0.000 \\
    \bottomrule
  \end{tabular}
\end{table}

\paragraph{Implications.}
Mid-band amplitude fidelity matters for stability: \emph{EI} exposes large magnitude distortions in GPT-4o/Llama/Mixtral, so downstream pipelines that depend on accurate scaling (e.g., compounding, normalization, controller gains) will drift unless corrected. DeepSeek’s best-in-class EI gain suggests safer use when amplitude tracking dominates, whereas Mixtral’s large RS error flags sensitivity to saturating transforms. Family-level selection thus changes which model is “best” for a given deployment.

% --- Mid-band |Phase Error| table ---
\begin{table}[h]
  \centering
  \small
  \caption{\textbf{A.2 Mean \(|\)Phase Error\(|\) (deg) at mid-frequencies (4 \& 8 cycles).} 
  \emph{Lower is better. LS is the timing bottleneck (largest lags/leads); Qwen is best on LS, while Mixtral collapses on RS.}}
  \label{tab:midband-phase}
  \begin{tabular}{lrrrrr}
    \toprule
    & \textbf{DeepSeek V3.1} & \textbf{GPT-4o} & \textbf{Llama 4 Instruct} & \textbf{Mixtral 8$\times$7B} & \textbf{Qwen3 235B} \\
    \midrule
    Exponential Interest & 4.47  & 0.24 & 3.54 & 0.04  & 2.97 \\
    Linear Solve         & 0.02  & 0.01 & 1.02 & 0.56  & 0.03 \\
    Linear System        & 26.38 & 7.38 & 4.49 & 42.40 & 2.61 \\
    Ratio Saturation     & 0.01  & 0.01 & 0.38 & 58.42 & 0.01 \\
    Similar Triangles    & 0.00  & 0.00 & 0.00 & 0.05  & 0.00 \\
    \bottomrule
  \end{tabular}
\end{table}

\paragraph{Implications.}
Phase governs \emph{timing consistency}: large LS phase errors (Mixtral, DeepSeek) imply lag/lead that can destabilize iterative procedures (solvers, rollouts) and corrupt ablations that assume time alignment. Qwen’s low LS phase is attractive for timing-sensitive use cases even if its gain is not always best. When choosing models for pipelines with feedback or chaining, prioritize low phase on the relevant family.

\begin{figure}[h]
  \centering
  \includegraphics[width=\linewidth]{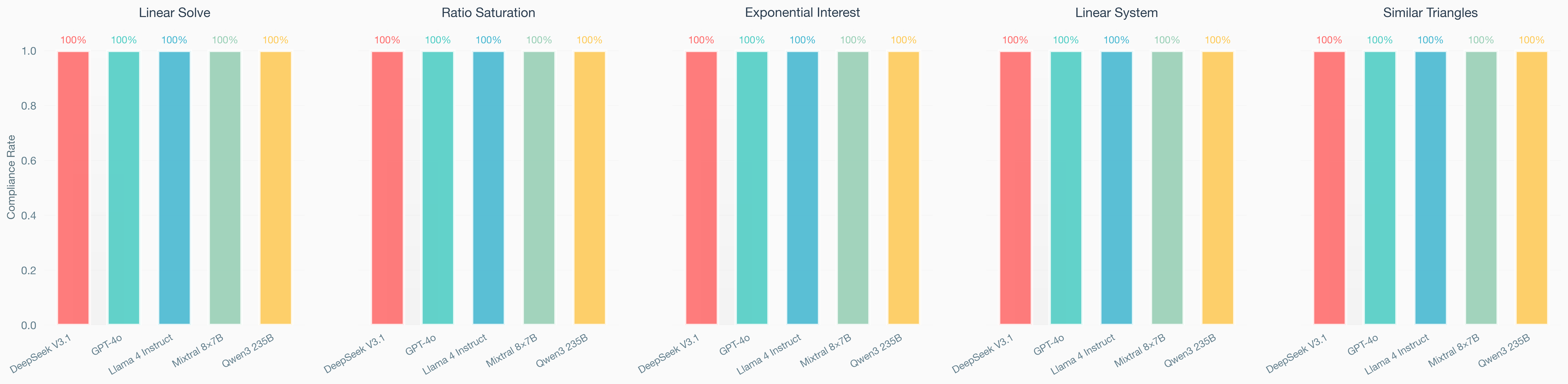}
  \caption{\textbf{A3. Compliance by family.} Compliance is perfect overall.}
  \label{fig:a2-compliance}
\end{figure}

\paragraph{Implications.}
Near-perfect compliance removes formatting as a confound: observed dynamics (gain/phase/residuals) reflect model behavior rather than parse failures. This also means MB-Plus penalties primarily capture quality, not I/O brittleness, and reproductions should match our curves given the same row IDs.

\begin{figure}[h]
  \centering
  \includegraphics[width=\linewidth]{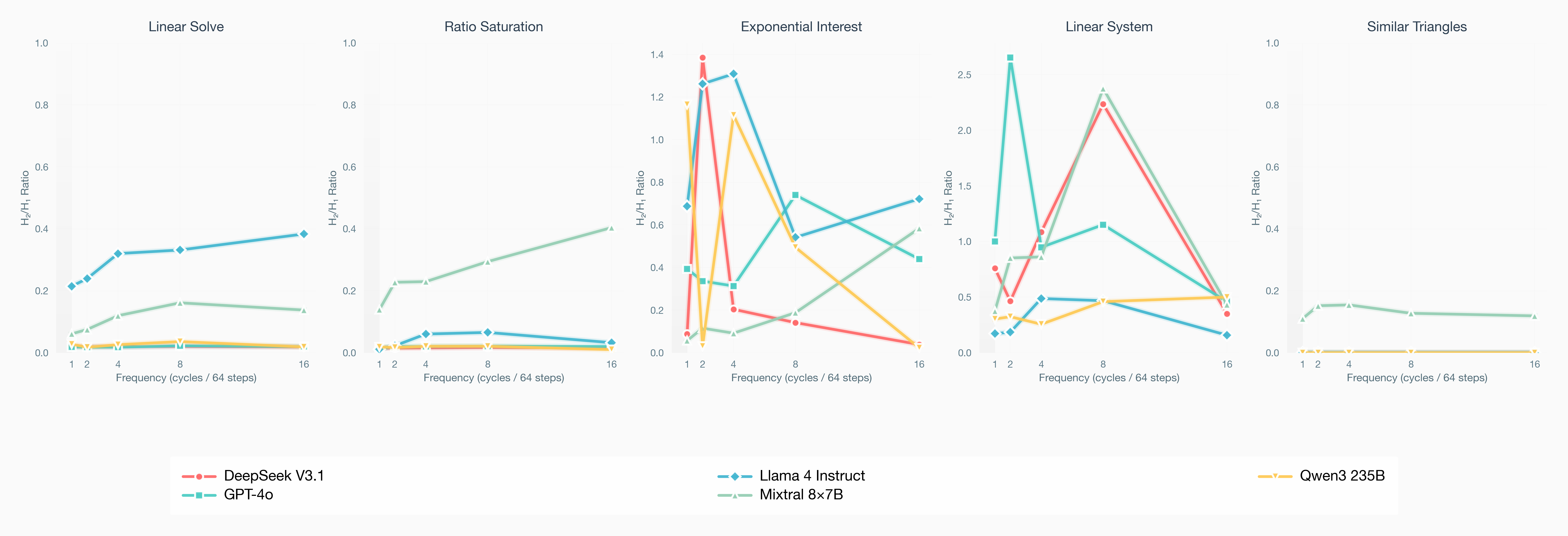}
  \caption{\textbf{A4. $H_2/H_1$ vs.\ frequency.} Nonlinearity concentrates in EI and LS; Similar Triangles stays near zero.}
  \label{fig:a3-h2h1}
\end{figure}

\paragraph{Implications.}
Elevated $H_2/H_1$ indicates distortion rather than pure linear gain/phase behavior. Peaks in EI/LS suggest that prompts with compounding or coupled relations will exhibit waveform deformation under parameter sweeps—use multi-tone tests or chirps to separate memory effects from static nonlinearity, and avoid using single-sinusoid fingerprints alone to claim linearity.

\begin{figure}[h]
  \centering
  \includegraphics[width=\linewidth]{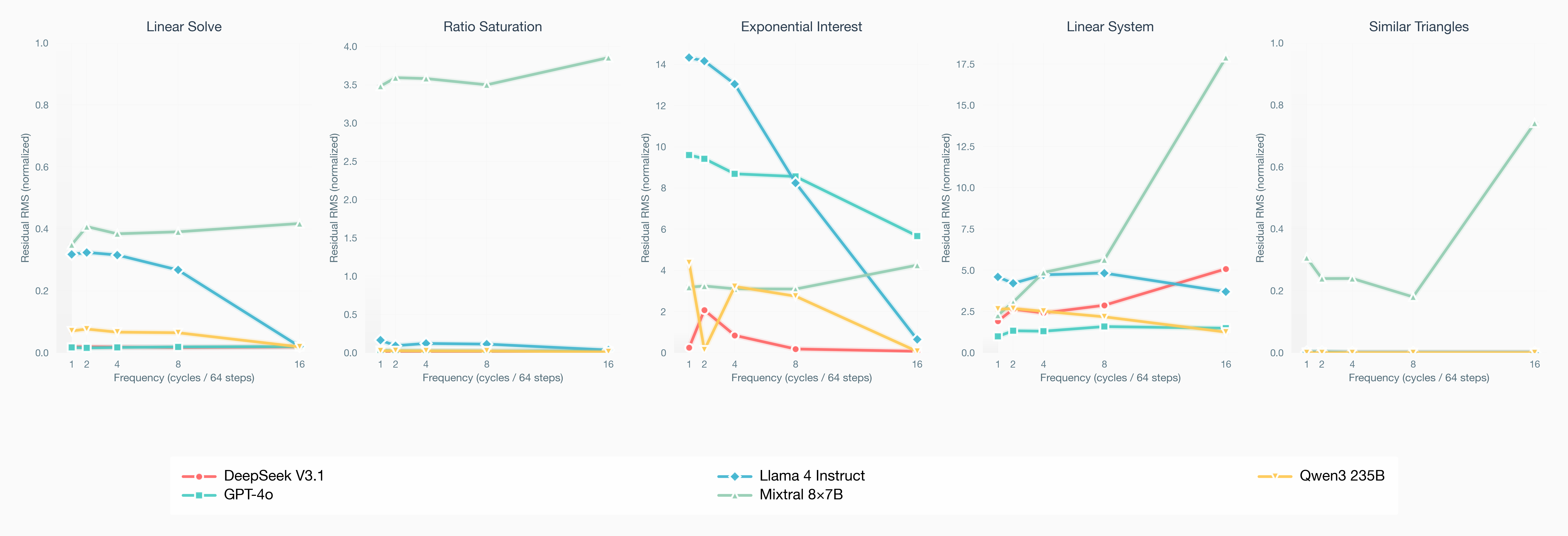}
  \caption{\textbf{A5. Residual RMS (normalized).} Single-sinusoid fits leave the largest residuals in EI and LS; simpler families fit tightly.}
  \label{fig:a4-resrms}
\end{figure}

\paragraph{Implications.}
High residuals mean a first-harmonic model is insufficient: EI/LS retain structure after removing the main tone, so downstream diagnostics should include richer inputs (chirps, steps, two-tone mixtures) before attributing errors solely to amplitude or timing. Low residuals on simpler families justify using mid-band summaries (MB-Core/MB-Plus) as compact, reliable proxies there.

\section{API Settings}
\label{app:api}

For all model calls (Together and OpenAI), we used the following fixed decoding settings:
\begin{itemize}
  \item \textbf{Temperature:} 0.0
  \item \textbf{Max tokens:} 1028
\end{itemize}

\noindent To ensure stable throughput and reproducibility, we applied simple rate limiters:
\begin{itemize}
  \item \textbf{Together:} 600 requests per minute (RPM)
  \item \textbf{OpenAI:} 20,000 tokens per minute (TPM)
\end{itemize}

\noindent These settings were held constant across all experiments unless explicitly noted elsewhere.

%%%%%%%%%%%%%%%%%%%%%%%%%%%%%%%%%%%%%%%%%%%%%%%%%%%%%%%%%%%%

\end{document}